\pdfoutput=1
\documentclass[11pt]{article}

\usepackage[preprint]{acl}
\usepackage{times}
\usepackage{latexsym}
\usepackage[T1]{fontenc}
\usepackage[utf8]{inputenc}
\usepackage{microtype}
\usepackage{inconsolata}
\usepackage{graphicx}
\usepackage{enumitem}
\usepackage{booktabs}
\usepackage{multirow}
\usepackage{adjustbox}
\usepackage{color}
\usepackage{xcolor}
\usepackage{tabularx}
\usepackage{subcaption}
\usepackage{hyperref}
\usepackage[most]{tcolorbox}
\usepackage{float}
\usepackage{inconsolata}

\title{Ontology-Free General-Domain Knowledge Graph-to-Text Generation Dataset Synthesis using Large Language Model}

\author{Daehee Kim$^1$, Deokhyung Kang$^1$, Sangwon Ryu$^1$ \textnormal{and} Gary Geunbae Lee$^1{^2}$ \\
  $^1$Graduate School of Artificial Intelligence, POSTECH, South Korea \\
  $^2$Department of Computer Science and Engineering, POSTECH, South Korea \\
  \texttt{\{andrea0119, deokhk, ryusangwon, gblee\}@postech.ac.kr} \\}

\begin{document}
\maketitle
\begin{abstract}
Knowledge Graph-to-Text (G2T) generation involves verbalizing structured knowledge graphs into natural language text. 
Recent advancements in Pretrained Language Models (PLMs) have improved G2T performance, but their effectiveness depends on datasets with precise graph-text alignment. 
However, the scarcity of high-quality, general-domain G2T generation datasets restricts progress in the general-domain G2T generation research. 
To address this issue, we introduce \textbf{Wiki}pedia \textbf{O}ntology-\textbf{F}ree \textbf{Graph}-text dataset (\textbf{WikiOFGraph}), a new large-scale G2T dataset generated using a novel method that leverages Large Language Model (LLM) and Data-QuestEval. 
Our new dataset, which contains 5.85M general-domain graph-text pairs, offers high graph-text consistency without relying on external ontologies. 
Experimental results demonstrate that PLM fine-tuned on WikiOFGraph outperforms those trained on other datasets across various evaluation metrics. 
Our method proves to be a scalable and effective solution for generating high-quality G2T data, significantly advancing the field of G2T generation.\footnote{Our code and data are available at: \url{https://github.com/daehuikim/WikiOFGraph}}
\end{abstract}

\section{Introduction}
\begin{figure}[t]
  \includegraphics[width=\columnwidth]{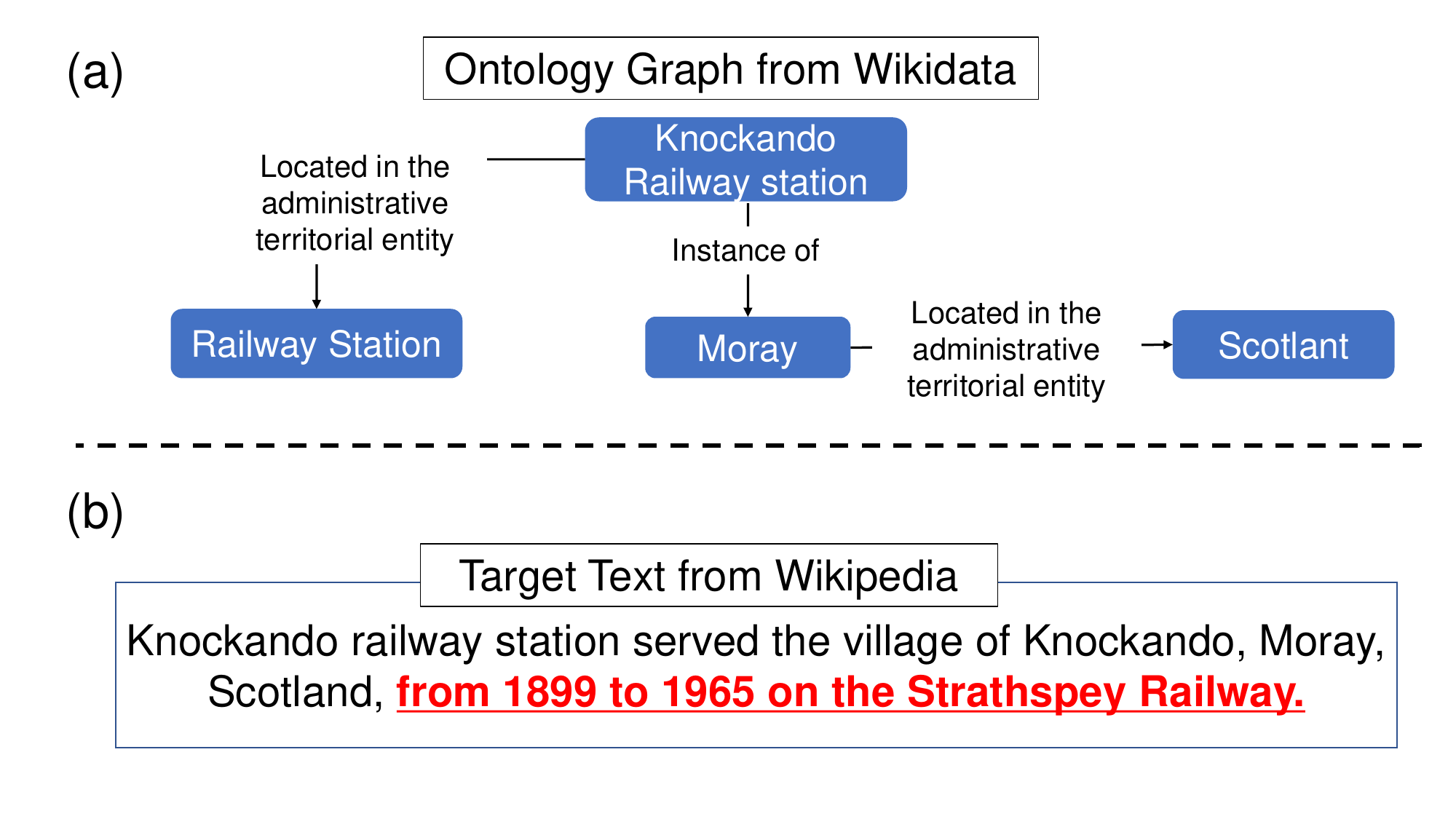}
  \caption{An example of a Graph-Text pair from existing ontology-based datasets. Although (a) and (b) are paired, the graph in (a) does not contain the information of the \textcolor{red}{\underline{underlined text}} in (b), illustrating a common misalignment problem.}
  \label{fig:1}
\end{figure} 
Knowledge Graph-to-Text Generation (G2T) is a task aimed at verbalizing knowledge graphs represented as a set of triplets in the form of \textit{(subject, predicate, object)} into natural language text \cite{10.5555/3241691.3241693,10215344}. 
Recent advancements in Pretrained Language Models (PLMs), such as T5 \cite{2020t5} and BART \cite{lewis-etal-2020-bart} have led to significant improvement in G2T when fine-tuned on G2T datasets \cite{kale-rastogi-2020-text}. 

However, fine-tuning PLMs requires a substantial amount of well-aligned G2T data \cite{kasner-etal-2023-mind}. 
Constructing G2T datasets is labor-intensive and expensive because understanding structured graph representations for various natural language sentences is challenging for human annotators.
Therefore, previous research has mainly focused on small-scale domain-specific datasets \cite{gardent-etal-2017-webnlg,koncel-kedziorski-etal-2019-text,castro-ferreira-etal-2020-2020,nan-etal-2021-dart}. 

To address this issue, researchers have attempted to automatically align Wikipedia texts with their corresponding graphs from ontologies to construct large-scale, general-domain G2T datasets \cite{elsahar-etal-2018-rex,jin-etal-2020-genwiki,agarwal-etal-2021-knowledge,wang-etal-2021-wikigraphs,mousavi-etal-2024-construction-paired}. 
However, ontology-based datasets often suffer from graph-text misalignment, making it challenging to generate text accurately.
For example, in Figure \ref{fig:1} (b), the phrase \textit{"from 1899 to 1965 on the Strathspey railway"} cannot be inferred solely from the graph shown in Figure \ref{fig:1} (a).
Such discrepancy can be a major cause of the decline in G2T performance \cite{mousavi-etal-2024-construction-paired}. 

Recent advances in Large Language Models (LLMs) offer promising potential to address these challenges.
Many studies have successfully leveraged LLMs to synthesize high-quality data for various tasks \cite{long-etal-2024-llms}.
Several studies have attempted to generate graph-text paired data using LLMs \cite{josifoski-etal-2023-exploiting,han-etal-2024-pive,chen-etal-2024-sac}, but these attempts have yet to be thoroughly explored for the G2T generation task. 

To address this issue, we introduce an effective method for generating high-quality G2T dataset that integrates LLM with Data-QuestEval \cite{rebuffel-etal-2021-data}. 
We select three examples from the human-crafted dataset to facilitate In-Context Learning (ICL) for the LLM, enabling it to extract graph representations from the given sentences.
To ensure high consistency between the extracted graph representations and text, we utilize Data-QuestEval for data curation.
Through this method, we create a new dataset called \textbf{Wiki}pedia \textbf{O}ntology-\textbf{F}ree \textbf{Graph}-Text dataset (\textbf{WikiOFGraph}), a 5.85M G2T data that covers a broad range of Wikipedia articles without relying on ontologies. 

Through comprehensive analyses, we demonstrate that our dataset achieves graph-text consistency comparable to that of a fully human-crafted dataset while significantly surpassing other ontology-based general-domain G2T datasets.
Experimental results demonstrate that a PLM fine-tuned on WikiOFGraph outperforms those trained on existing datasets across the human expert-crafted GenWiki \cite{jin-etal-2020-genwiki} test set and LLM-synthesized WikiOFGraph test set. 
This highlights the suitability of our dataset for building G2T systems that perform well across general domains, making it more effective than other datasets.
We further demonstrate the effectiveness of Data-QuestEval filtering through additional experiments. \\
In summary, our key contributions are as follows:
\begin{enumerate}[label=(\roman*)]
\item We introduce an effective method for synthesizing high-quality G2T dataset.
Our approach is independent of proprietary LLMs or ontologies, making it reproducible and easily adaptable for various domains.
\item We release a new G2T generation dataset called \textbf{WikiOFGraph}.
Our comprehensive analyses indicate that it offers high graph-text consistency, comparable to a human-crafted dataset, with 5.85M samples covering the entire spectrum of Wikipedia.
\item We demonstrate the effectiveness of Data-QuestEval filtering through additional experiments and case study.
\end{enumerate}

\section{Background and Related Work}
\paragraph{Graph-to-Text generation}
Traditionally, researchers tackled G2T generation using templates designed to expressing predicates into pre-defined statements \cite{wiseman-etal-2018-learning,kasner-dusek-2022-neural,xiang-etal-2022-asdot,vejvar-fujimoto-2023-aspiro}.
Template-based approaches have low hallucination rates but are labor-intensive in creating various templates, and they often struggle with producing fluent sentences from complex graphs \cite{kasner-dusek-2020-data}. 

To address these limitations, researchers have leveraged neural encoder-decoder architectures \cite{wiseman-etal-2017-challenges,beck-etal-2018-graph,nie-etal-2018-operation,DBLP:conf/aaai/Puduppully0L19,iso-etal-2019-learning} which convert graph representations into vector embeddings interpretable by neural models.
Unlike previous approaches, these approaches leverage an end-to-end G2T generation paradigm that does not require pre-defined templates.

The introduction of transformer \cite{NIPS2017_3f5ee243}-based PLMs has significantly advanced G2T generation, leading to much better performance than earlier methods. 
\citeauthor{kale-rastogi-2020-text} demonstrated substantial improvements by fine-tuning these PLMs on G2T generation tasks. 
This strategy has since been adopted in subsequent studies \cite{ribeiro-etal-2021-investigating, article, mehta-etal-2022-improving, han-shareghi-2022-self}.
However, this new paradigm requires substantial amounts of well-aligned G2T data. 

\begin{figure*}[t]
\centering
\includegraphics[width=0.9\linewidth]{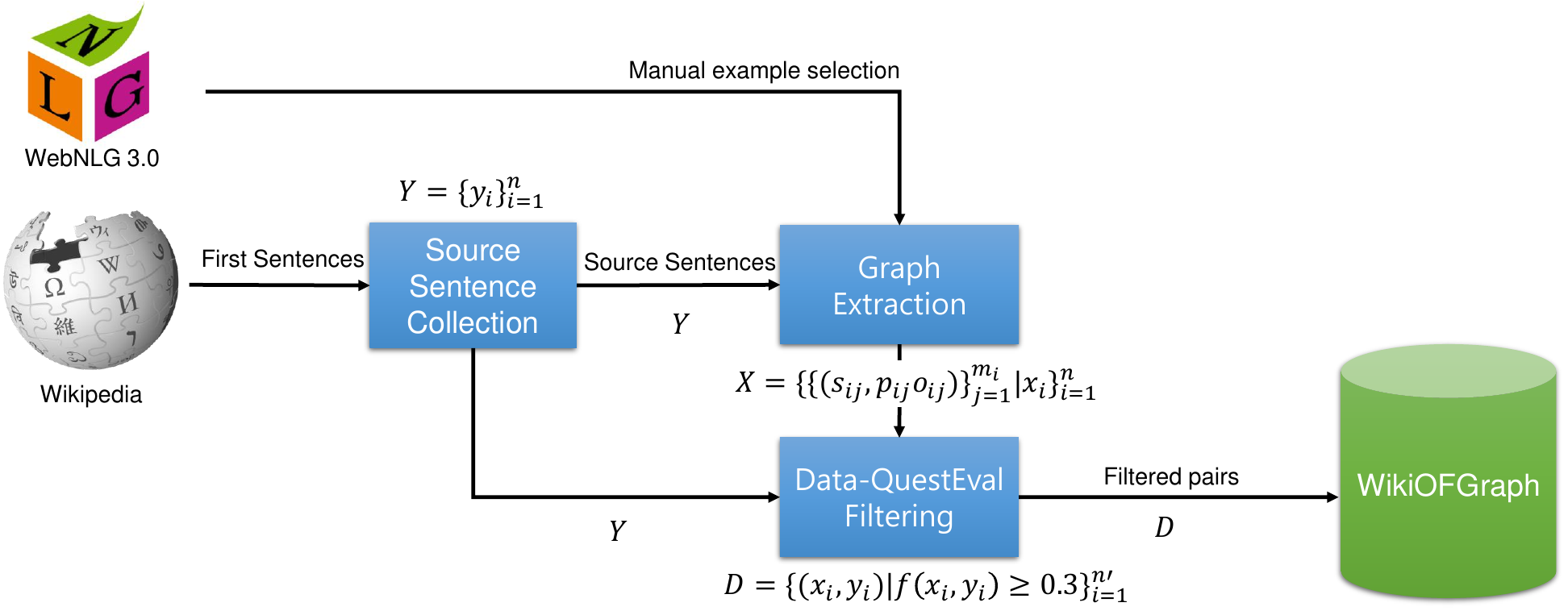} \hfill
  \caption {Method for constructing the \textbf{WikiOFGraph}. Source sentences are collected from Wikipedia. 
Graph representations are then extracted using an LLM through in-context learning, guided by manually selected examples from the WebNLG.
Data-QuestEval Filtering curates graph-text pairs compiled into the \textbf{WikiOFGraph}.}
  \label{fig:2}
\end{figure*}

\paragraph{G2T generation datasets}
WebNLG \cite{gardent-etal-2017-webnlg, castro-ferreira-etal-2020-2020} is a fully human-crafted dataset based on the DBpedia \cite{mendes-etal-2012-dbpedia}, making it a representative benchmark for G2T generation tasks due to its precise graph-text alignment. 
However, creating human-crafted datasets is resource-intensive and limits scale. 
These limitations on scale and diversity have led researchers to develop large-scale G2T datasets automatically.

GenWiki \cite{jin-etal-2020-genwiki} provides a large-scale, general-domain G2T dataset with 1.3 M samples.
It includes fine and full splits, determined by the F1-score-based alignment between text and graph entities, and 1,000 human-crafted test samples.
TekGen \cite{agarwal-etal-2021-knowledge} combines Wikipedia and Wikidata \cite{10.1145/2629489} to create a large-scale, general-domain G2T dataset using a distant supervision strategy.
This approach allowed for the creation of large datasets but led to insufficient alignment between graph representations and target texts.
LAGRANGE \cite{mousavi-etal-2024-construction-paired}, a 3.0M data, aligns Wikidata and Wikipedia data and incorporates second-hop reasoning and triplet augmentation.

However, these attempts rely on text aligning algorithms based on ontologies with insufficient information, leading to inadequate graph-text alignment. 
In contrast, our method generates graphs directly from the source text, better capturing the text's content. 
Our method is simple and scalable, enabling the creation of significantly larger datasets with greater domain diversity than existing ontology-based datasets.

\paragraph{LLM-driven data generation}
The natural language processing community has recently shifted its focus toward synthesizing high-quality, large-scale data using LLMs \cite{long-etal-2024-llms}. 
While LLM-generated data is easily applicable to discriminative tasks \cite{whitehouse-etal-2023-llm,ghosh-etal-2024-abex}, its effectiveness in generative tasks diminishes when the distribution of the generated data differs from existing data \cite{ding2024dataaugmentationusinglarge}.

While several studies have explored using LLMs for generating graph-text pairs, their primary focus has been on tasks such as Text-to-Graph (T2G) generation rather than on G2T generation \cite{josifoski-etal-2023-exploiting,han-etal-2024-pive,mousavi-etal-2024-construction-paired,chen-etal-2024-sac}. 
In contrast, our research focuses on the LLM-driven synthetic graph generation for creating G2T generation dataset.
Moreover, our approach leverages Data-QuestEval for data filtering, ensuring high graph-text consistency without manual or heuristic adjustments.

\section{WikiOFGraph}
In this section, we introduce a method for creating \textbf{WikiOFGraph}. 
We first define the requirements necessary to address the limitations of existing G2T datasets (§\hyperref[sec:31]{3.1}).
We then outline the rule-based algorithm for gathering source sentences from Wikipedia (§\hyperref[sec:32]{3.2}), as well as the process of extracting graphs through LLM (§\hyperref[sec:33]{3.3}).
Additionally, we describe our approach to utilizing Data-QuestEval \cite{rebuffel-etal-2021-data} for curating well-aligned graph-text pairs (§\hyperref[sec:34]{3.4}).
\subsection{Requirements}
\label{sec:31}
\paragraph{Graph-text consistency} Good G2T data guarantees that the target text includes \textit{"all"} and \textit{"only"} information from the graph.
If \textit{"all"} information of the graph is not reflected in the target text, it can cause data omission problems for the model. 
If the target text does not includes \textit{"only"} information from the graph, it can cause hallucination problems for the model.
Therefore, graph-text consistency is essential to prevent omission and hallucination problems in the G2T generation task.
\paragraph{Domain diversity}
PLMs fine-tuned on datasets limited to a specific domain often struggle to handle samples from unseen domains \cite{keymanesh-etal-2022-makes}. 
This issue becomes apparent when PLMs are tested on new domains, where performance falls short compared to familiar domains \cite{castro-ferreira-etal-2020-2020}. 
Therefore, including a wide range of domains is crucial to developing a domain-adaptive G2T generation system. 
\paragraph{Large scale}
The demand for substantial structured datasets is growing with the emergence of language models trained on extremely large-scale structured data \cite{zhuang2024structlm,li-etal-2024-unifying}.
High-quality structured data, such as knowledge graph representations, is essential for training LLMs for structured data processing.
However, creating such data is far more challenging than compiling unstructured corpora. 
Consequently, providing a large-scale G2T dataset could be key to advancing LLMs in handling structured data effectively.

\subsection{Source Sentences Collection}
\label{sec:32}
To meet the requirements (§\hyperref[sec:31]{3.1}), we select Wikipedia as the source of sentences. 
Wikipedia includes sentences with extensive factual knowledge from diverse domains.
We employ a rule-based algorithm to extract the first sentence from every article on English Wikipedia because the first sentence of a Wikipedia article often encapsulates vital factual information.
We apply several constraints to enhance the clarity of the factual context in the source sentences.  
We limit the length of sentences to 10-500 characters, exclude sentences starting with pronouns to avoid ambiguous entity extraction (e.g., 'it,' 'that'), and remove parenthetical explanations to prevent redundant expressions.
Through this process, we collect 6.06M source sentences, denoted as $Y = \{y_i\}_{i=1}^n$ in Figure \ref{fig:2}, where $n$ represents the total number of source sentences.

\subsection{Graph Extraction}
\label{sec:33}
We utilize the \textit{"Llama-3-70b-instruct-awq"}\footnote{Available at: \href{https://huggingface.co/casperhansen/llama-3-70b-instruct-awq}{casperhansen/llama-3-70b-instruct-awq}} \cite{dubey2024llama3herdmodels} to extract graph representations directly from $Y$.
We manually select three examples from the WebNLG \cite{castro-ferreira-etal-2020-2020} dataset for in-context examples \footnote{For implementation details, see the \hyperref[appendix:graph]{Appendix A}}. 
We then obtain a set of graph representations for each sentence, denoted as $X = \left\{\left\{\left(s_{ij}, p_{ij}, o_{ij}\right)\right\}_{j=1}^{m_i} \mid x_i\right\}_{i=1}^{n}$ in Figure \ref{fig:2}.  
Here, $n$ represents the total number of sentences, $m_i$ denotes the number of triplets associated with the $i$-th sentence, and $(s_{ij}, p_{ij}, o_{ij})$ are the subject, predicate, and object of the $j$-th triplet in the $i$-th sentence, respectively.
Through this process, we obtain 6.06M pairs of source sentences and their corresponding generated graph representations.

\subsection{Data-QuestEval Filtering}
\label{sec:34}
Data-QuestEval \cite{rebuffel-etal-2021-data} is a framework that measures the accuracy of predicted texts in Data-to-Text generation tasks by combining question generation and question-answering techniques to evaluate whether the \textit{predicted sentence} accurately reflects the content of the \textit{source data}.
Data-QuestEval supports referenceless evaluation by relying solely on the \textit{source data} and the \textit{predicted sentence}. 
We adapt referenceless evaluation by using the $X$ as the \textit{source data} and the $Y$ as the \textit{predicted sentence}.

We apply the referenceless Data-QuestEval scoring $f(x_i, y_i)$ to evaluate the consistency between the generated graph representations and their corresponding source sentences. Only pairs with a score of $f(x_i, y_i) \geq 0.3$ are retained for further curation, as denoted by $D = \{(x_i, y_i) \mid f(x_i, y_i) \geq 0.3\}_{i=1}^{n'}$ in Figure \ref{fig:2}, where $n'$ represents the total number of curated graph-text pairs.
Through this process, we obtain 5.95M\footnote{We employ 5.85M samples for the training split and 100K samples for the test split.} samples, with less than 2\% of the samples being filtered out.
This result implies that a robust LLM can reliably extract graph representations containing \textit{"all"} and \textit{"only"} information from given texts.

\section{Data Analysis}
\label{sec:4}
We compare \textbf{WikiOFGraph} with existing datasets. 
We explore the scale and domain diversity of the data through quantitative analysis (§\hyperref[sec:41]{4.1}).
We then assess the graph-text consistency using human evaluators and GPT-4o \cite{openai2024gpt4technicalreport} (§\hyperref[sec:42]{4.2}).

We utilize four representative G2T datasets for our comparative analysis: human-crafted WebNLG \cite{castro-ferreira-etal-2020-2020} and three automatically generated ontology-based datasets—GenWiki\footnote{We employ the GenWiki\textsubscript{FINE} training split} \cite{jin-etal-2020-genwiki}, TekGen \cite{agarwal-etal-2021-knowledge}, and LAGRANGE \cite{mousavi-etal-2024-construction-paired}. Details of the datasets are provided in the appendix (§\hyperref[appendix:data]{Appendix D}).

\subsection{Quantitative Analysis}
\label{sec:41}
\begin{table}[H]
\centering
\resizebox{\columnwidth}{!}{%
\begin{tabular}{@{}ccccc@{}}
\toprule
                  Dataset & \# samples & \# unique predicate & \# unique entity & \# triplet (m/M/avg)
                   \\ \midrule
WebNLG             & 35K        & 372                 & 3.2K             & 1/7/2.96
\\
GenWiki            & 680K       & 287                 & 86.6K            & 1/10/2.64
\\
TekGen             & \textbf{6.31M}      & 50,861              & 4.3M            & 1/54/1.73
\\
LAGRANGE           & 3.07M      & 1,167               & 2.9M             & 1/135/4.02
\\ \midrule
WIkiOFGraph (Ours) & 5.85M      & \textbf{140,733}    & \textbf{8.2M}   & 1/17/3.62           
\\ \bottomrule
\end{tabular}%
}
\caption{Training set statistics for comparative analysis. \textit{\# triplet (m/M/avg)} indicates the minimum, maximum, and average number of triplets per sample.}
\label{tab:1} 
\end{table}
\paragraph{Domain diversity \& dataset volume} 
Table \ref{tab:1} provides a detailed comparison of statistics across various G2T datasets.
WikiOFGraph includes a remarkably higher number of unique predicates and unique entities than other datasets. 
This diversity in predicates and entities is anticipated to enhance domain generalization in G2T generation tasks.
In addition, WikiOFGraph is advantageous in terms of scale, containing a substantial 5.85 M data pairs, which makes it a valuable resource for future research in G2T generation tasks.

In contrast, WebNLG contains the smallest number of unique predicates, entities, and samples. 
This limitation is primarily due to its entirely human-crafted nature, which makes it challenging to produce a large volume of data. 

GenWiki and LAGRANGE also have fewer unique predicates and entities.
This is because the underlying ontology graphs in these datasets cover a relatively narrow scope. 
Also, these datasets are designed to support G2T and T2G generation tasks, limiting the variety of available predicate expressions.

\begin{figure}[H]
    \centering
    \includegraphics[width=0.9\linewidth]{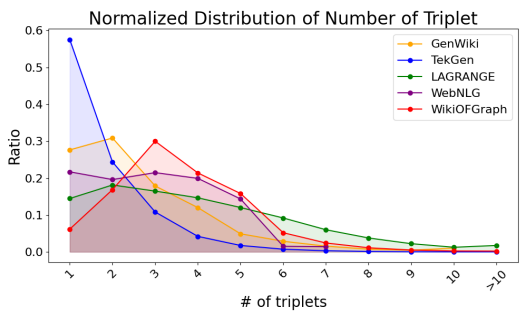}
    \caption{Normalized distribution of the number of triplets in each dataset.}
    \label{fig:3}
\end{figure}
\begin{figure}[H]
    \centering
    \includegraphics[width=0.9\linewidth]{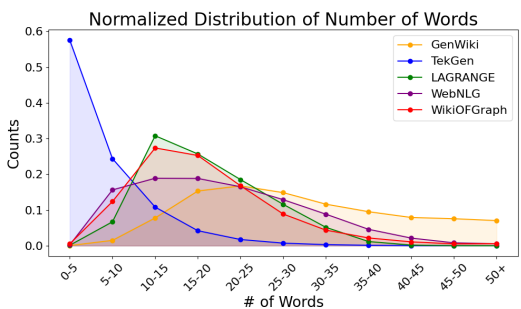}
    \caption{Normalized distribution of the number of words in each dataset.}
    \label{fig:4}
\end{figure}
\paragraph{Statistical comparison of dataset compositions} To provide more detailed comparisons, we conduct a statistical analysis of these datasets. 
Figure \ref{fig:3} shows the distribution of samples categorized by the number of triplets, while Figure \ref{fig:4} presents the distribution of samples categorized by word count in increments of five.
WikiOFGraph demonstrates a balanced and consistent dataset structure, with the distribution of triplets and word counts closely aligning.
The similarity in the shapes of these two distributions, which differs from those of other datasets, suggests a consistent alignment between the graph and the target text.

Despite TekGen showing good alignment between triplet and word count distributions, most of its samples consist of two or fewer triplets or contain fewer than ten words, indicating that most samples are relatively short.
GenWiki predominantly consists of samples with three or fewer triplets; however, the relatively high proportion of samples with more than 20 words suggests a surface-level inconsistency between the graph representations and the corresponding texts.
\begin{figure}[H]
    \centering
    \includegraphics[width=\linewidth]{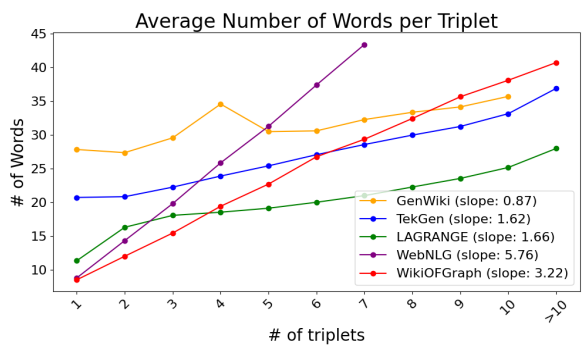}
    \caption{Average number of words per number of triplets across different datasets.}
    \label{fig:5}
\end{figure}
\paragraph{Average words per triplet} We also visualize the average number of words per sample for each dataset, categorized by the number of triplets, as shown in Figure \ref{fig:5}.
WikiOFGraph exhibits an average word increase rate of 3.22 per triplet, which closely aligns with the average of 3.62 triplets per sample, as reported in Table \ref{tab:1}. 
This suggests that WikiOFGraph provides consistent information with minimal bias relative to the number of triplets.

WebNLG exhibits a relatively high word increase rate of 5.76 per triplet compared to other datasets. 
However, this rate contrasts with the average of 2.96 triplets per sample, as reported in Table 1. 
This discrepancy arises because most WebNLG samples consist of short sentences and few triplets.

On the other hand, the automatically generated ontology-based datasets show much lower word increase rates per triplet, ranging from 0.87 to 1.66. 
These lower rates suggest that each triplet in these datasets is associated with fewer words, leading to less detailed representations per triplet.

\subsection{Qualitative Analysis}
\label{sec:42}
To further analyze graph-text consistency, we conduct a qualitative analysis employing human and GPT-4o. 
We provide 30 samples per dataset, each evaluated by five human reviewers who assess the same set of samples, and an additional 1,000 samples per dataset are evaluated by GPT-4o.
Since some datasets have a high ratio of samples with a low number of triplets, we categorize the samples by the number of triplets and randomly sample them within each range to ensure a diverse evaluation across different sample lengths.

Our focus is to assess whether the information of graph representations (triplets) are accurately reflected in the target text. 
To achieve this, we ask evaluators to review graph-text pairs and identify 1) unused triplets that are not used to generate the text and 2) parts of the text that could not be guessed from the triplets.
\begin{figure}[H]
    \centering
    \includegraphics[width=\linewidth]{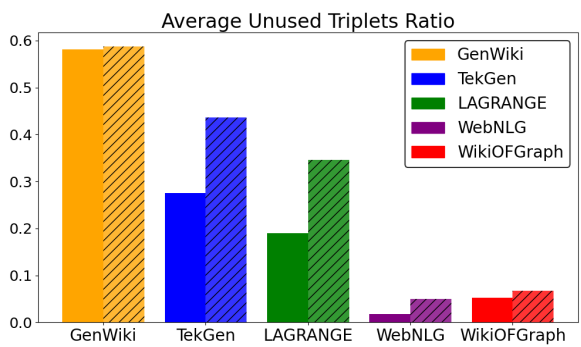}
    \caption{Number of unused triplet ratio}
    \label{fig:6}
\end{figure}
\begin{figure}[H]
    \centering
    \includegraphics[width=\linewidth]{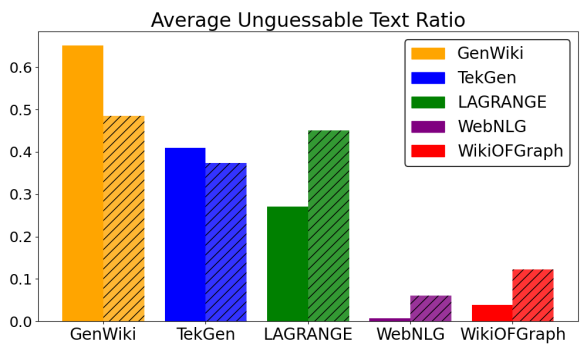}
    \caption{Length of unguessable text ratio}
    \label{fig:7}
\end{figure}
\paragraph{Graph-text consistency} Figure \ref{fig:6} presents the average ratio of unused triplets across different datasets. 
The plain bars represent the evaluations conducted by human evaluators, while the hatched bars indicate the assessment results from GPT-4o.
When comparing automatically generated datasets, the average ratio of unused triplets is highest in GenWiki, followed by TekGen, LAGRANGE, and WikiOFGraph.
Notably, WikiOFGraph is comparable to WebNLG, a fully human-crafted dataset.
This observation suggests that WikiOFGraph incorporates triplet information into the text better than automaticaly generated ontology-based datasets.

Figure \ref{fig:7} presents the average ratio of unguessable text across different datasets. 
Similarly, GenWiki, TekGen, and LAGRANGE exhibit relatively high amounts of unguessable text, indicating a significant presence of text that cannot be generated from the graph information alone. 
In contrast, WikiOFGraph contains a low amount of unguessable text, even comparable to WebNLG. 
This observation suggests that WikiOFGraph has significantly higher graph-text consistency than automaticaly generated ontology-based G2T datasets. 
Details regarding the evaluation process can be found in the appendix (§\hyperref[appendix:qa]{Appendix B}).
\begin{table*}[ht]
\centering
\begin{adjustbox}{max width=\textwidth}
\begin{tabular}{cccccccc}
\toprule
                   & BLEU ($\uparrow$)  & ChrF++ ($\uparrow$)           & METEOR ($\uparrow$)         & TER ($\downarrow$)           & BLEURT ($\uparrow$)       & ROUGE-L ($\uparrow$)      & BertScore-F1 ($\uparrow$)     \\ \midrule
WebNLG             & 35.97   & 65.76        & 39.69          & 49.71          & 0.44          & 64.41         & 95.56          \\ \hline
GenWiki            & 36.14    & 56.96 & 33.80     &  89.77        & 0.13         &  58.13             & 92.85         \\
TekGen             &  39.25    &  63.43  &  38.19     &  75.15   & 0.30         &    64.16           &  94.13       \\
LAGRANGE           & 40.40     & 67.51        & 40.70           & 47.15          & 0.43          & 65.26         & 95.71          \\ \midrule
WikiOFGraph (Ours) & \textbf{45.85}   & \textbf{69.61} &  \textbf{42.17} & \textbf{43.38} & 0.\textbf{46} & \textbf{70.10} & \textbf{95.89} \\ 

\bottomrule
\end{tabular}
\end{adjustbox}
\caption{Evaluation result on GenWiki test set.}
\label{tab:2}
\end{table*}

\begin{table*}[ht]
\centering
\begin{adjustbox}{max width=\textwidth}
\begin{tabular}{cccccccc}
\toprule
                   & BLEU ($\uparrow$)  & ChrF++ ($\uparrow$)           & METEOR ($\uparrow$)         & TER ($\downarrow$)           & BLEURT ($\uparrow$)       & ROUGE-L ($\uparrow$)      & BertScore-F1 ($\uparrow$)     \\ \midrule
WebNLG             & 21.91   & 35.97        & 39.69          & 49.71          & 0.44          & 62.88         & 93.81         \\ \hline
GenWiki            & 23.42    & 48.30  &  33.17     &  74.89       & 0.14   & 61.92    & 92.14      \\
TekGen             &  38.45    & 62.28    & 41.39   & 55.01      &  0.45    & 73.08  & 94.62    \\
LAGRANGE           & 39.37     & 63.33        & 42.85           & 46.04          & 0.51          & 74.68         & 95.08          \\ \midrule
WikiOFGraph (Ours) & \textbf{69.27}   & \textbf{82.63} & \textbf{51.59} & \textbf{17.44} & \textbf{0.71} & \textbf{86.46} & \textbf{98.29} \\ \bottomrule
\end{tabular}
\end{adjustbox}
\caption{Evaluation result on WikiOFGraph test setin.}
\label{tab:3}
\end{table*}

\section{Experiments}
\label{sec:5}
We provide a comprehensive overview of our experimental setup and results.
We first outline our experimental settings (§\hyperref[sec:51]{5.1}). 
We then describe the evaluation metrics used to assess the model's G2T generation performance (§\hyperref[sec:52]{5.2}). 
We describe the evaluation data selection, focusing on ensuring a broad topic range and high reliability (§\hyperref[sec:53]{5.3}). 

\subsection{Experimental Settings}
\label{sec:51}
We choose the T5-large \cite{2020t5}, which has 770M parameters, for our backbone model. 
The backbone model is trained to take \textit{source triplets} as input and predict a \textit{target text}. 
We use special tokens such as '<S>', '<P>', and '<O>' to distinguish triplet elements. 
For example, a \textit{source triplets} like \textit{"(<S> Arròs negre| <P> country| <O> Spain), (<S> Spain| <P> ethnic Group| <O> Spaniards)"} is used to predict a \textit{target text} such as \textit{"Arros negre is from Spain where Spaniards are an ethnic group."} 
We utilize Huggingface's transformers \cite{wolf-etal-2020-transformers} library for our experiments.
We use the cross-entropy loss function during training.
Implementation details and hyperparameter settings are provided in the appendix (§\hyperref[appendix:ft]{Appendix C}).

\subsection{Evaluation Metrics}
\label{sec:52}
We employ representative metrics commonly used to evaluate the performance of G2T generation tasks.
We utilize several metrics, including BLEU \cite{papineni-etal-2002-bleu}, METEOR \cite{banerjee-lavie-2005-meteor}, ChrF++ \cite{popovic-2015-chrf}, TER \cite{snover-etal-2006-study}, BLEURT \cite{sellam-etal-2020-bleurt} and ROUGE-L \cite{lin-2004-rouge}, specifically to evaluate the fluency of the predicted sentences. 
To assess semantic accuracy, we employ BERTScore-F1 \cite{bert-score}.

\subsection{Evaluation Data Selection}
\label{sec:53}
We utilize evaluation datasets that cover a broad range of topics while maintaining high reliability to assess the model's general domain G2T capabilities.
First, we choose the GenWiki\cite{jin-etal-2020-genwiki} test set, consisting of 1,000 samples generated by human experts, for its high level of trustworthiness. 
Next, we select 100,000 samples from the WikiOFGraph dataset not included in the training split as an additional evaluation dataset. 
Although WikiOFGraph consists of synthesized data, our earlier analyses indicate that it possesses high domain diversity and strong graph-text consistency. 
This makes it an appropriate choice for evaluating the model's G2T generation capabilities in the general domain.

\section{Results and Discussions}
\label{sec:6}
Table \ref{tab:2} and \ref{tab:3} present the performances of the T5-large model fine-tuned on five different datasets, with the best values highlighted in bold.
The results demonstrate that fine-tuning with WikiOFGraph consistently achieves the highest performance across various metrics on both test sets.

Training on domain-specific, human-crafted datasets like WebNLG leads to lower performance in general domain G2T generation tasks due to the limited size and narrow coverage of the training data. 
On the other hand, training on datasets that cover a broader range of domains but are automatically generated based on ontology also leads to lower performance due to reduced graph-to-text consistency.
This highlights the importance of dataset scale, domain diversity, and graph-text consistency in achieving robust performance in general domain G2T generation as mentioned in §\hyperref[sec:31]{3.1}.

Table \ref{tab:3} highlights a more apparent performance gap between the T5-large model fine-tuned on WikiOFGraph and those fine-tuned on other datasets.
This performance gap is anticipated due to the similar distribution of test samples to WikiOFGraph, yet the difference is considerably significant.
These results demonstrate that WikiOFGraph is significantly more effective than other G2T datasets for general domain G2T generation.

\begin{table*}[ht]
\centering
\begin{adjustbox}{max width=\textwidth}
\begin{tabular}{@{}cccc@{}}
\toprule
Status&Graph Representations  & Text                                                                                                                       & Data-QuestEval Score \\ \midrule
Incomplete Text Error&(\textless{}S\textgreater Piano sonata no.| \textless{}P\textgreater Type| \textless{}O\textgreater Music composition)       & \textcolor{red}{The Piano Sonata No.} & 0.1123 \\ \midrule
Incomplete Text Error&(\textless{}S\textgreater Dübs crane tank| \textless{}P\textgreater Model| \textless{}O\textgreater No.)       & \textcolor{red}{Dübs crane tank No.} & 0.1097 \\ \midrule
Ambiguous Pronoun Error&(\textless{}S\textgreater Mnl-2| \textless{}P\textgreater Round| \textless{}O\textgreater Qualification round)  & \textcolor{red}{This was} the qualification round for MNL-2. & 0.1270  \\ \midrule
Ambiguous Pronoun Error&\begin{tabular}[c]{@{}c@{}}(\textcolor{red}{\textless{}S\textgreater S}| \textless{}P\textgreater Name| \textcolor{red}{\textless{}O\textgreater O})\\ (\textcolor{red}{\textless{}S\textgreater S}| \textless{}P\textgreater Operator| \textcolor{red}{\textless{}O\textgreater O})\end{tabular}                                                                     & \textcolor{red}{This is} the name of an aerial lift, as well as its operator.                                                               & 0.1142 \\ \midrule
Properly Generated&(\textless{}S\textgreater Domenico Puccini| \textless{}P\textgreater studied Under| \textless{}O\textgreater Giovanni Paisiello)       & Domenico Puccini studied for a time under Giovanni Paisiello. & 0.4999 \\ \midrule

Properly Generated&\begin{tabular}[c]{@{}c@{}}(\textless{}S\textgreater Dennis Hamilton| \textless{}P\textgreater signed Data| \textless{}O\textgreater October 21, 1967)\\ (\textless{}S\textgreater Dennis Hamilton| \textless{}P\textgreater signed By| \textless{}O\textgreater Los Angeles Lakers)\end{tabular} & \begin{tabular}[c]{@{}c@{}}October 21, 1967 The Los Angelos Lakers signed \\ Dennis Hamilton as a free agent.\end{tabular} & 0.6768 \\ \midrule
Properly Generated&(\textless{}S\textgreater Double Hill Station| \textless{}P\textgreater location| \textless{}O\textgreater up the Rakaia River)       & Double Hill Station, located up the Rakaia River. & 0.8929 \\ 

\bottomrule
\end{tabular}%
\end{adjustbox}
\caption{Examples of Graph-Text pairs with corresponding Data-QuestEval Scores, ranging from 0 to 1. Text highlighted in \textcolor{red}{red} indicates potential errors or areas of concern.}
\label{tab:5}
\end{table*}

\section{Effectiveness of Data-QuestEval}
\label{sec:7}
Due to the extremely low ratio of filtered samples (less than 2\%), observing significant performance differences from Data-QuestEval curation is challenging.
To further investigate the impact of Data-QuestEval, we conduct additional experiments with controlled sample sizes (§\hyperref[sec:71]{7.1}). 
We then conduct a case study to verify whether Data-QuestEval effectively excludes misaligned graph-text pairs (§\hyperref[sec:72]{7.2}).

\subsection{Impact of Filtered Sample Ratios}
\label{sec:71}

\begin{table}[H]
\centering
\resizebox{\columnwidth}{!}{%
\begin{tabular}{@{}ccccc@{}}
\toprule
Curated:Noise & BLEU ($\uparrow$)       & METEOR ($\uparrow$)        & Rouge-L ($\uparrow$)       & BertScore-F1 ($\uparrow$)  \\ \midrule
0:100         & 43.21       & 40.62          & 66.11          & 95.56       \\
25:75         & 44.25       & 41.01          & 67.47          & 95.79       \\
50:50         & 44.57       & 41.22          & 67.88          & 95.77       \\
75:25         & 44.75       & 41.16          & 68.38          & 95.83       \\
100:0         & \textbf{45.00} & \textbf{41.37} & \textbf{69.32} & \textbf{96.00} \\ \bottomrule
\end{tabular}%
}
\caption{Impact of \textit{Curated:Noise} sample ratios on model performance.}
\label{tab:4}
\end{table}
We define \textit{'curated'} samples as those scoring 0.3 or higher in the Data-QuestEval filtering, while samples scoring below 0.3  are classified as \textit{'noise'}. 
We randomly select these \textit{'curated'} samples from WikiOFGraph, while the \textit{'noise'} samples are from the filtered-out samples.
Given WikiOFGraph's large scale, increasing the ratio of noise directly in the dataset is challenging.
Therefore, we fix the train split at 50,000 samples and vary the ratio of \textit{'curated'} to \textit{'noise'} samples. 
The experimental settings used for these experiments follows the procedure outlined in §\hyperref[sec:51]{5.1}.

As shown in Table \ref{tab:4}, the evaluation result of the fine-tuned T5-large model improves proportionally as the ratio of curated samples increases.
Notably, all key metrics—including BLEU, METEOR, Rouge-L, and BertScore-F1—exhibit consistent improvement, indicating a comprehensive enhancement in model performance.
This result indicates that our approach to measuring graph-text consistency through Data-QuestEval filtering is effective and highly practical, particularly in scenarios with limited access to high-quality training data.

\subsection{Case Study}
\label{sec:72}
We conduct a case study to understand the low graph-text consistency problems in samples filtered-out by Data-QuestEval. 
Through the case study, we recognize that the filtered-out samples exhibit two representative types of problems.
We highlight these error types with examples along with their corresponding Data-QuestEval scores in Table \ref{tab:5}.

The first error type arises from incomplete source sentences. 
For instance, a sentence like \textit{"Piano Sonata No. 1, the default title for a composer's first (or only) piano sonata, may refer to:"} can be split as an incomplete sentence such as \textit{"Piano Sonata No."} due to its punctuation mark, which causes confusion in the \textit{sentence splitter}.
These incomplete sentences make it difficult to extract graph representations accurately.

The second error type occurs due to the use of ambiguous demonstratives or pronouns in the source sentence.
Sentences containing demonstratives like \textit{"This"} or other ambiguous pronouns make it challenging to determine the subject-object structure accurately.

Lastly, we observe that most well-structured sentences, like \textit{"October 21, 1967 The Los Angeles Lakers signed Dennis Hamilton as a free agent."}, generally result in consistent graph representations.

These results highlight the importance of using complete and unambiguous sentences for graph extraction while demonstrating that Data-QuestEval filtering effectively identifies \textit{'noise'} samples from the G2T dataset. 

\section{Conclusion}
This study introduces WikiOFGraph, a large-scale G2T generation dataset with 5.85M samples covering the entire Wikipedia domain.
To address the issue of graph-text misalignment commonly found in ontology-based datasets, we propose a novel method leveraging LLM and Data-QuestEval to generate high-quality graph-text pairs.
Comprehensive analyses reveal that WikiOFGraph achieves \textbf{graph-text consistency} comparable to the fully human-crafted dataset, while also exhibiting \textbf{large scale} and extensive \textbf{domain diversity}.
Comparisons with representative G2T datasets demonstrate that fine-tuning PLMs with WikiOFGraph significantly enhances their ability to perform general domain G2T tasks.
Additional experiments and a case study demonstrate the effectiveness of Data-QuestEval in ensuring high-quality graph-text alignments, reinforcing its value in data curation. 
Our approach provides a scalable and efficient method for generating high-quality G2T data without relying on proprietary LLMs, external ontologies or extensive human involvement, making it reproducible for advancing G2T generation.

\section*{Acknowledgements}
This work was supported by Smart HealthCare Program(www.kipot.or.kr) funded by the Korean National Police Agency(KNPA, Korea) [Project Name: Development of an Intelligent Big Data Integrated Platform for Police Officers’ Personalized Healthcare / Project Number: 220222M01].

This research was supported by the MSIT(Ministry of Science and ICT), Korea, under the ITRC(Information Technology Research Center) support program(IITP-2024-2020-0-01789) supervised by the IITP(Institute for Information \& Communications Technology Planning \& Evaluation).

\section*{Limitations}
\paragraph{Multilingual extension}
Although our method is easily reproducible across various domains using open-source LLMs and public source texts, the scope of our study focuses mainly on English.
Future research could extend the study's multilingual capabilities by employing LLMs with multilingual capabilities and utilizing source texts from multilingual corpora. 
\paragraph{Room for improvement}
We apply various empirical rules of thumb, particularly in areas such as prompt engineering and sampling strategies.
While our approach significantly outperforms existing datasets in meeting key requirements, there remains room for improvement, particularly in refining prompts, optimizing sampling parameters for graph extraction, and selecting the most effective LLM. 
Additionally, since Data-QuestEval might not be the most optimal measure, exploring alternative methods to assess the consistency between the generated graph representations and the source text could be a valuable direction for future work.
\paragraph{Expanding directions}
Our work focuses primarily on the direction of G2T generation. 
Building a dataset that also supports T2G generation poses challenges, such as the need to select predicate expressions from a predefined vocabulary, which could limit flexibility. 
Future work could address this by reassigning predicate expressions within the WikiOFGraph dataset, potentially creating a more adaptable framework for T2G tasks.
\paragraph{Data contamination concerns}
One important limitation not yet addressed in our study is the potential for data contamination. 
Since the LLMs used in our experiments are pre-trained on publicly accessible datasets like Wikipedia, they may already be familiar with the data used in our graph extraction tasks.
To thoroughly assess whether our method is robust against data contamination, future research should replicate our experiments on a large corpus that is not part of the LLM's pre-training data.

\section*{Ethical Considerations}
We prioritize transparency and reproducibility in our research by using widely accessible resources.
We utilize the open-source LLM \textit{Llama3-70b-instruct-awq}\footnote{\textbf{WikiOFGraph} is built with Meta Llama3; Meta Llama 3 is licensed under the Meta Llama 3 Community License, Copyright © Meta Platforms, Inc. All Rights Reserved. More details can be found at: \href{https://llama.meta.com/llama3/license/}{Llama3 License}} and the publicly available Wikipedia\footnote{Wikipedia is licensed under the Creative Commons Attribution-ShareAlike 3.0 Unported License (CC BY-SA 3.0). The specific terms of this license can be found at:  \href{https://en.wikipedia.org/wiki/Wikipedia:Text_of_the_Creative_Commons_Attribution-ShareAlike_3.0_Unported_License}{Wikipedia License}} dataset, adhering to their respective licenses.

For human evaluation, we conduct the assessments through Upwork\footnote{\href{https://www.upwork.com}{https://www.upwork.com}}, ensuring participants receive appropriate compensation for their efforts. 
Each evaluator is paid a fixed rate of \$60 for their work, which involves evaluating five datasets for approximately three hours. 
This compensation is determined to be fair and in line with industry standards, reflecting our commitment to ethical research practices and the fair treatment of participants.

\bibliography{custom}

\appendix
\section{Details of Graph Extraction} 
\label{appendix:graph}
We utilize the vLLM \cite{10.1145/3600006.3613165} for efficient inference using GPUs with at least 40 GB of VRAM. 
We perform graph extraction using nucleus sampling \cite{DBLP:conf/iclr/HoltzmanBDFC20} based decoding strategy with a temperature set to 0.5 and a top-p value of 0.9, which are determined experimentally.
Total inference time for processing 6M data samples was approximately 600 hours when utilizing 4 GPUs in parallel.
Therefore, we divided the data into divisions that contain 100K samples each to generate the outputs in parallel.
Detailed prompts are provided in Table \ref{tab:8}.

\section{Details of  Qualitative Analysis}
\label{appendix:qa}
We describe details of human evaluation process and results (§\hyperref[appendix:human]{B.1}). We then describe GPT-4o evaluation process and prompts (§\hyperref[appendix:gpt]{B.2}). 

\subsection{Details of human evaluation} 
\label{appendix:human}

\begin{table}[H]
\centering
\resizebox{\columnwidth}{!}{%
\begin{tabular}{@{}ccccccc@{}}
\toprule
                   &         & GenWiki & TEKGEN & LAGRANGE & WebNLG & WikiOFGraph \\ \midrule
\multirow{2}{*}{1} & Triplet &  63.52       &  33.26      &  24.42        &  2.25      &  7.45           \\
                   & Text    & 78.38        & 56.66       & 37.88         &  1.26      &  4.61           \\ \midrule
\multirow{2}{*}{2} & Triplet & 67.19        & 32.80       & 16.78         &  2.38      &  4.29           \\
                   & Text    &  72.83       &  13.61      &  5.03        &  0.96      &  3.33           \\ \midrule
\multirow{2}{*}{3} & Triplet & 43.00        &  20.77      &  10.29        &  0.00      &  2.78           \\
                   & Text    &  71.80       &  47.59      & 28.26         & 0.23       &  6.27           \\ \midrule
\multirow{2}{*}{4} & Triplet &   47.28      &  17.46      &  30.71        &  4.16      &  10.00           \\
                   & Text    & 67.61        &  34.65      & 28.85         & 0.80       &  2.78           \\ \midrule
\multirow{2}{*}{5} & Triplet &  69.63       &  33.39      &  12.84        & 0.00       & 1.67            \\
                   & Text    &  35.00       & 52.45       &   35.39       &  0.25      & 2.52            \\ \midrule
\multirow{2}{*}{Average} & Triplet &   58.12±12.14      &    27.54±7.78    &     19.01±8.44     &    1.76±1.77    &   5.24±3.44          \\
                   & Text    &  65.12±17.27       &   40.99±17.4     &    27.08±13      &    0.7±0.45    & 3.9±1.55            \\ \bottomrule
                   
\end{tabular}%
}
\caption{Individual scores of 5 different human evaluators. The \textit{Average} row includes the mean and standard deviation in the form of \textit{mean±standard deviation}}
\label{tab:6}
\end{table}

\begin{figure*}[t]
\centering
\includegraphics[width=0.9\linewidth]{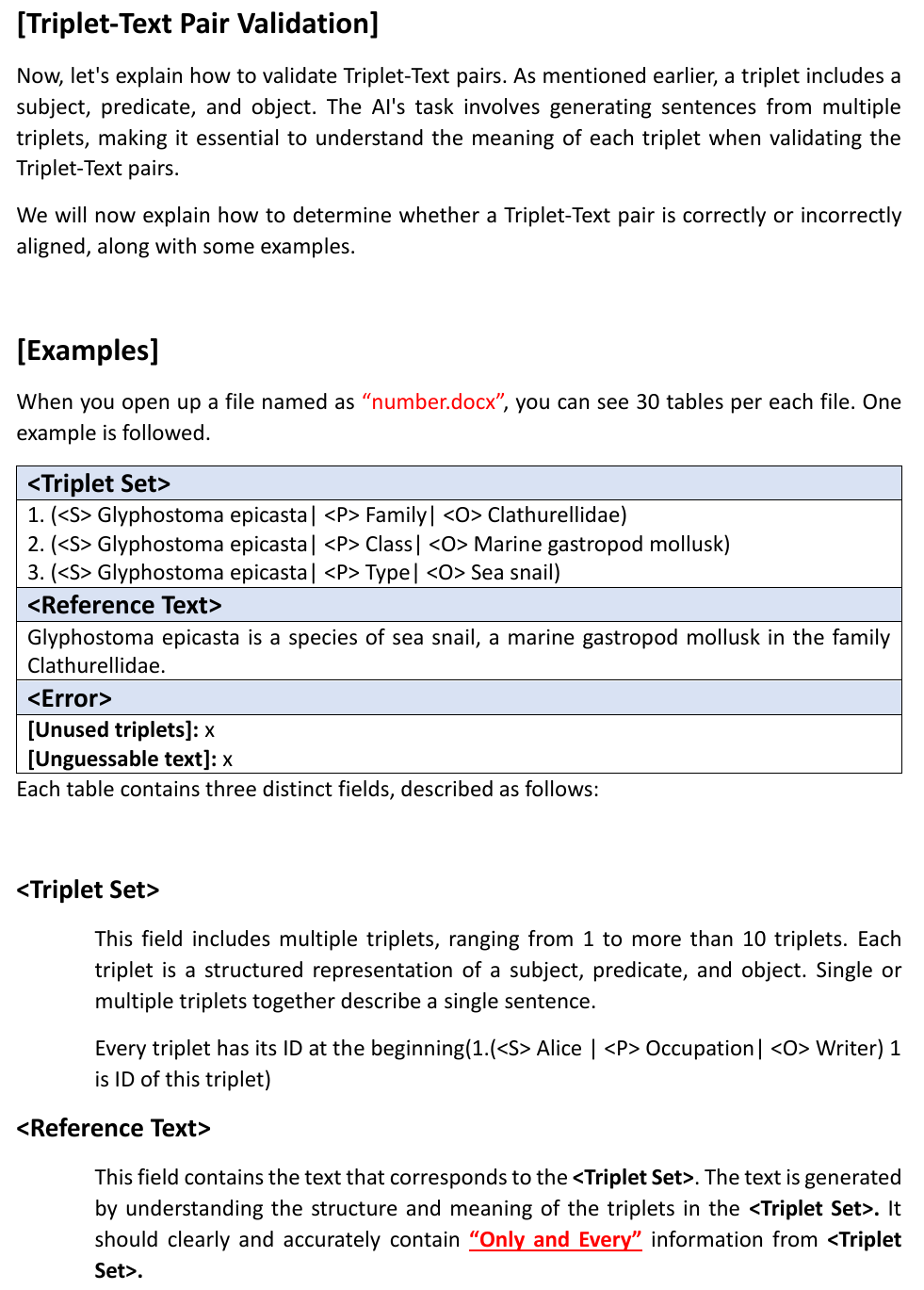} \hfill
  \caption {Part of the guidelines for human evaluators.}
  \label{fig:8}
\end{figure*}

\begin{figure*}[t]
\centering
\includegraphics[width=0.9\linewidth]{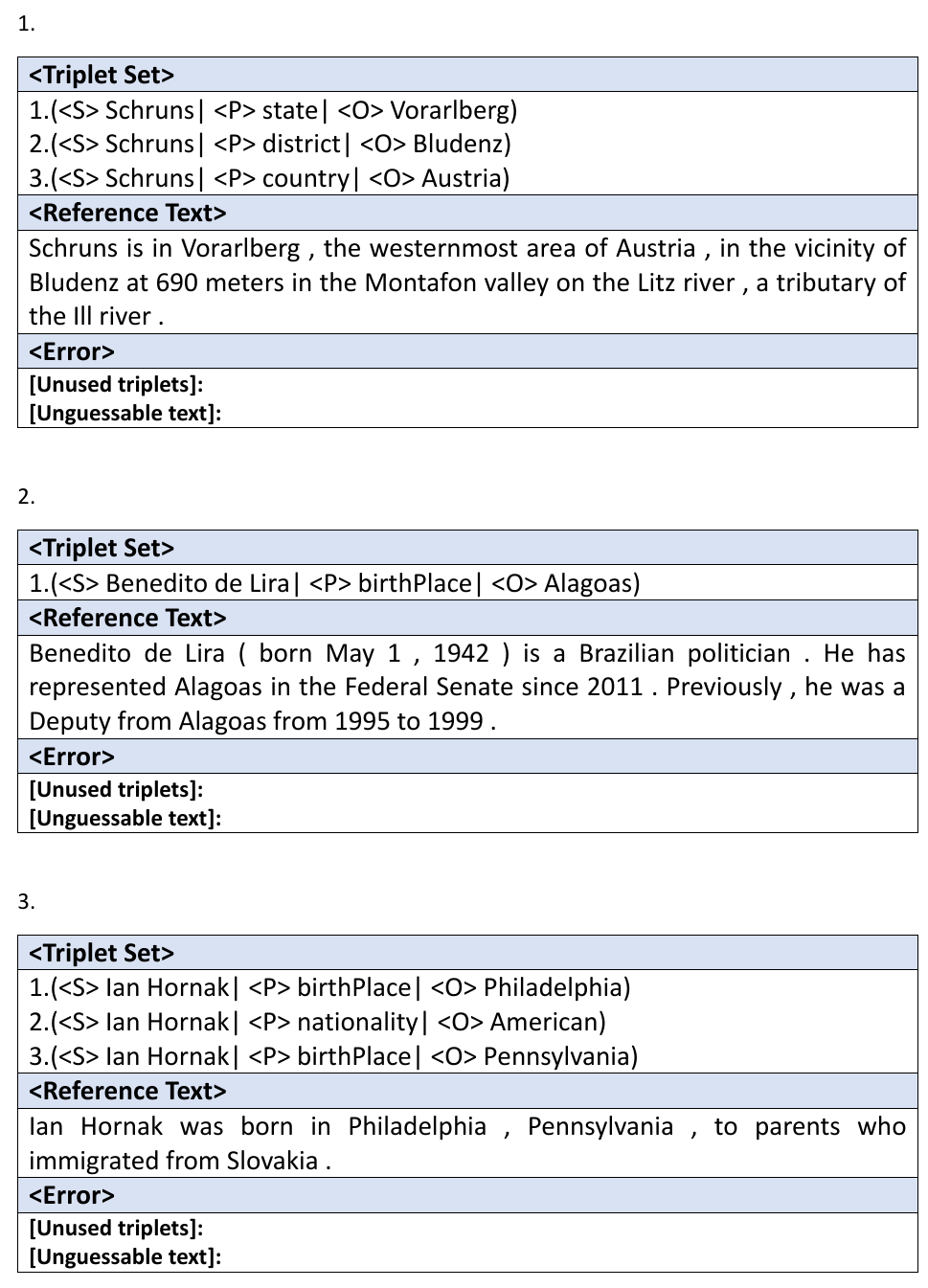} \hfill
  \caption {Some examples of human evaluation tasks.}
  \label{fig:9}
\end{figure*}

The human evaluators consist of a diverse group of five individuals from various professional backgrounds. 
All evaluators, with their native English proficiency or advanced English language certifications, have prior experience in AI data labeling. 
We provide the human evaluators with an overview of the guideline and examples to validate graph-text pairs. 
Evaluators are tasked with assessing whether the target text accurately reflects the information contained within the corresponding triplets.
Part of the evaluation guideline\footnote{The full guideline is available at \url{https://github.com/daehuikim/WikiOFGraph}.} provided with the human evaluators is shown in Figure \ref{fig:8}.
We give detailed examples and step-by-step instructions to ensure consistency and accuracy to the human evaluators. 
Some examples of an actual evaluation task completed by the evaluators are shown in Figure \ref{fig:9}.
Following this process, we collect responses from five human evaluators. 
The aggregated results, showing the response results for each evaluator, are presented in Table \ref{tab:6}.

\subsection{Details of GPT-4o evaluation}
\label{appendix:gpt}
For the GPT-4o evaluation, we craft a prompt that closely mirrored the guideline provided to the human evaluators, ensuring that the LLM could comprehend and process the task effectively. 
Using the \textit{gpt-4o-2024-05-13 model}, we evaluated 1,000 samples across five different datasets, resulting in a total cost of \$39.04.
The actual prompt used in this evaluation can be found in Table \ref{tab:9}.
We set the temperature to 1 and allowed it to generate up to 700 tokens.
The results to assess the consistency of 5 different datasets are shown in Table \ref{tab:7}.

\begin{table}[H]
\centering
\resizebox{\columnwidth}{!}{%
\begin{tabular}{@{}cccccc@{}}
\toprule
        & GenWiki & TEKGEN & LAGRANGE & WebNLG & WikiOFGraph \\ \midrule
Triplet &  58.74       & 43.62       &  34.64        &  4.93      &  6.71           \\
Text    &  48.56       &  37.37      &  45.06        & 6.03       &  12.23           \\ \bottomrule
\end{tabular}%
}
\caption{Scores of GPT-4o evaluation.}
\label{tab:7}
\end{table}

\section{Details of Fine-Tuning} 
\label{appendix:ft}
In the section \hyperref[sec:5]{5}, we apply bf16 precision, a cosine learning rate decay function with 2k warmup steps, and utilize ZeRO stage 3 optimization through DeepSpeed \cite{10.1145/3394486.3406703} while employing the AdamW optimizer \cite{DBLP:conf/iclr/LoshchilovH19}. 
We employ beam search decoding for training and inference, with a temperature set to 1 and beam size set to 5, to ensure a controlled search space exploration while maintaining the diversity of generated outputs.
The specific hyperparameters vary depending on the dataset to ensure adequate optimization steps.

For WebNLG \cite{castro-ferreira-etal-2020-2020}, we set the gradient accumulation steps and the number of parallel GPUs to achieve an effective batch size of 8 with a maximum learning rate of 0.3. 
We evaluate every 4K steps and selected the model with the lowest loss. 
Total training time was approximately 13 hours using 4 GPUs in parallel (52 GPU hours).

For other datasets, we use an effective batch size of 192 with a maximum learning rate of 0.5. 
Evaluations are performed every 20k steps, and the model with the minimum loss is selected. 
Total training time was approximately 70 hours using 4 GPUs in parallel (280 GPU hours).

In the experiments described in section \hyperref[sec:7]{7}, we limit the number of data samples to 50K, similar to the size of WebNLG. 
Therefore, we conduct the experiments using the same hyperparameter settings as those applied to WebNLG.
We select the model based on the one with the lowest loss as determined by evaluations conducted every 4K steps.
Total training time was approximately 10 hours using 4 GPUs in parallel (40 GPU hours).

\section{Details of datasets}
\label{appendix:data}
In this section, we provide information on the sources of the datasets used in our study and detailed statistics for each dataset.
\begin{enumerate}
\item \textbf{Wikipedia:} We use Wikipedia from Huggingface datasets. Specifically, we utilize the \textit{20220301.en} split, which can be accessed at \href{https://huggingface.co/datasets/legacy-datasets/wikipedia}{Wikipedia dataset}.

\item \textbf{WebNLG:} We use WebNLG obtained from the official repository. We utilize the English data from version 3.0, which is available at \href{https://gitlab.com/shimorina/webnlg-dataset}{WebNLG repository}.

\item \textbf{GenWiki:} We obtain GenWiki from \href{https://github.com/zhijing-jin/genwiki}{GenWiki repository}. We use the data found in the `fine` split. Since there is no separate evaluation split, we reserve 10\% of the training samples for evaluation purposes.

\item \textbf{TekGen:} We obtain TekGen from \href{https://github.com/google-research-datasets/KELM-corpus}{Tekgen repository}. We convert the data into triplets according to the JSON object rules specified in the official repository.

\item \textbf{LAGRANGE:} We obtain LAGRANGE from the footnotes of the paper \cite{mousavi-etal-2024-construction-paired}. 

\end{enumerate}
We also provide a detailed comparison of the specific statistics and characteristics of each dataset, offering a clear overview of their unique features and differences in Table \ref{tab:data}.

\begin{table*}[t]
\centering
\begin{tabular}{|p{0.95\textwidth}|}
\hline
\textbf{Instruction} \\ 
Your task is to create a set of triplets that can represent all the entities that appear in the given text. \\ 
Triplet is consist of three parts (\textless S\textgreater, \textless P\textgreater, \textless O\textgreater). \\ 
\textless S\textgreater\ means subject, \textless P\textgreater\ means predicate (relation), \textless O\textgreater\ means object. \\ 
You can not simply copy the entities from the text, you need to create a set of triplets that can represent all the entities in the text. \\ 
For example, you cannot just use "is" or "are" in \textless P\textgreater\ part, you need to find a more specific predicate that can represent the relationship between the subject and the object. \\ 
Complete the \textlbrackdbl TRIPLET\textrbrackdbl\ to represent \textlbrackdbl TEXT\textrbrackdbl, as shown in the Examples. \\ 
Please just complete \textlbrackdbl TRIPLET\textrbrackdbl\ without saying anything else. \\ \\
\textbf{Example 1} \\ 
\textlbrackdbl TEXT\textrbrackdbl: The Acharya Institute of Technology is located in Soldevanahalli, Acharya Dr. Sarvapalli Radhakrishnan Road, Hessarghatta Main Road, Bangalore – 560090, Karnataka, India. It is affiliated with the Visvesvaraya Technological University in Belgaum. \\ 
\textlbrackdbl TRIPLETS\textrbrackdbl: (\textless S\textgreater Acharya Institute of Technology| \textless P\textgreater affiliation| \textless O\textgreater Visvesvaraya Technological University), (\textless S\textgreater Visvesvaraya Technological University| \textless P\textgreater city| \textless O\textgreater Belgaum), (\textless S\textgreater Acharya Institute of Technology| \textless P\textgreater state| \textless O\textgreater Karnataka), (\textless S\textgreater Acharya Institute of Technology| \textless P\textgreater country| \textless O\textgreater India), (\textless S\textgreater Acharya Institute of Technology| \textless P\textgreater campus| \textless O\textgreater In Soldevanahalli, Acharya Dr. Sarvapalli Radhakrishnan Road, Hessarghatta Main Road, Bangalore – 560090.) \\ 
\textbf{Example 2} \\ 
\textlbrackdbl TEXT\textrbrackdbl: Albert Jennings Fountain was born in Staten Island in New York City and died in Dona Ana County, New Mexico. \\ 
\textlbrackdbl TRIPLETS\textrbrackdbl: (\textless S\textgreater Albert Jennings Fountain| \textless P\textgreater death Place| \textless O\textgreater Doña Ana County, New Mexico), (\textless S\textgreater Albert Jennings Fountain| \textless P\textgreater birth Place| \textless O\textgreater New York City), (\textless S\textgreater Albert Jennings Fountain| \textless P\textgreater birth Place| \textless O\textgreater Staten Island) \\ 
\textbf{Example 3} \\ 
\textlbrackdbl TEXT\textrbrackdbl: Abilene, Texas is in Jones County in the United States. Washington, D.C. is the capital of the U.S. with New York City being the largest city. English is their native language. \\ 
\textlbrackdbl TRIPLETS\textrbrackdbl: (\textless S\textgreater United States| \textless P\textgreater capital| \textless O\textgreater Washington, D.C.), (\textless S\textgreater Abilene, Texas| \textless P\textgreater is Part Of| \textless O\textgreater Jones County, Texas), (\textless S\textgreater Jones County, Texas| \textless P\textgreater country| \textless O\textgreater United States), (\textless S\textgreater United States| \textless P\textgreater largest City| \textless O\textgreater New York City), (\textless S\textgreater United States| \textless P\textgreater language| \textless O\textgreater English language) \\ \\
\textbf{Query} \\ 
\textlbrackdbl TEXT\textrbrackdbl: \textcolor{blue}{\{text\}} \\ 
\textlbrackdbl TRIPLETS\textrbrackdbl: \\ \hline
\end{tabular}
\caption{Actual prompt used for Graph Extraction. The text in \textcolor{blue}{blue} indicates where we have inserted the source sentences.}
\label{tab:8}
\end{table*}

\begin{table*}[t]
\centering
\begin{tabular}{|p{0.95\textwidth}|}
\hline
\textbf{Instruction} \\ 
Your job is to validate Triplet-Text pair data for Triplet-to-Text generation task. Based on the provided instructions and examples, write all errors in the Triplet-Text pairs present in the given query. \\ \\

\textbf{Triplet Composition:} \\
Format: (\textless S\textgreater subject\textbar \textless P\textgreater predicate\textbar \textless O\textgreater object) \\ 
A triplet is composed of three elements: subject, predicate, and object. Each of these elements is delineated by specific symbols with capital letters such as \textless S\textgreater, \textless P\textgreater, and \textless O\textgreater. These symbols help to clearly identify the role of each element within the triplet. \\
\textbullet\ \textless S\textgreater\ stands for the subject, which represents the entity that is performing an action or being described. \\
\textbullet\ \textless P\textgreater\ stands for the predicate, which illustrates the relationship or action that connects the subject to the object. \\
\textbullet\ \textless O\textgreater\ stands for the object, which is the entity that is receiving the action or being described in relation to the subject. \\ \\

\textbf{Error Types:} \\
A. Unused Triplet \\
\{Example\} \\
\{Explanation\} \\
B. Unguessable Text \\
\{Example\} \\
\{Explanation\} \\

Please just fill the format below. No need to write any extra explanations. \\ \\

\textbf{Query} \\ 
\textlbrackdbl TRIPLET SET\textrbrackdbl: \textcolor{blue}{\{triplet\}} \\ 
\textlbrackdbl TEXT\textrbrackdbl: \textcolor{blue}{\{text\}} \\ 
\textlbrackdbl ERRORS\textrbrackdbl: \\ 
\textbullet\ [Unused Triplets]:  \\
\textbullet\ [Unguessable Text]:  \\ \hline
\end{tabular}
\caption{Actual prompt for validating triplet-text pairs. \{Example\} and \{Explanation\} are omitted due to the length of the prompt. The text in \textcolor{blue}{blue} indicates where we have inserted the examples.}
\label{tab:9}
\end{table*}

\begin{table*}
\centering
\begin{adjustbox}{max width = \textwidth}
\begin{tabular}{@{}cccccc@{}}
\toprule
\multicolumn{1}{c}{}    & WebNLG        & TekGen              & GenWiki                   & LAGRANGE                                                                           & WIkiOFGraph (Ours) \\ \midrule
\# of samples           & 35,426        & \textbf{6,310,061}  & 681,436                   & 3,075,058                                                                          & 5,851,776          \\
\# of unique predicate  & 372           & 50,861              & 287                       & 1,167                                                                              & \textbf{140,733}   \\
\# of unique entity     & 3,211         & 4,249,337           & 866,373                   & 2,904,407                                                                          & \textbf{8,217,819} \\
\# of triples (m/M/avg) & 1/7/2.96      & 1/54/1.73           & 1/10/2.64                 & 1/135/4.02                                                                         & 1/17/3.62          \\
\midrule
Human Annotator         & o             & x                   & x                         & x                                                                                  & x                  \\
\midrule
Domain Diversity         & $\downarrow$             & $\uparrow$                   & $\uparrow$                         & $\uparrow$                                                                                  & \textbf{$\uparrow$}                  \\
\midrule
Chracteristic           & Crowdsourcing & Distant supervision & For unsupervised learning & \begin{tabular}[c]{@{}c@{}}Second-hop matching,\\ Triple Augmentation\end{tabular} & LLM graph extraction     \\
\midrule
Ontology                & DBPedia       & Wikidata            & DBPedia                   & Wikidata                                                                           & x                  \\ \bottomrule
\end{tabular}
\end{adjustbox}
\caption{Detailed comparison of dataset statistics and characteristics.}
\label{tab:data}
\end{table*}
\end{document}